\documentclass[11pt,a4paper]{article}
\usepackage[hyperref]{acl2021}
\usepackage{times}
\usepackage{latexsym}

\usepackage{microtype}

\usepackage{graphicx}
\usepackage{amsmath}
\usepackage{amsthm}
\usepackage{booktabs}
\usepackage[ruled,linesnumbered]{algorithm2e}
\usepackage{multirow}
\usepackage{amssymb}
\usepackage{color}
\usepackage{multirow}
\usepackage{natbib}
\usepackage{array}
\usepackage{graphicx}
\usepackage{bbm}
\usepackage{float}

\aclfinalcopy 

\title{Locate and Label: A Two-stage Identifier for Nested Named Entity Recognition}

\author{
\textbf{Yongliang Shen$^{1}$, Xinyin Ma$^{1}$, Zeqi Tan$^{1}$, Shuai Zhang$^{1}$, Wen Wang$^{2}$}, \textbf{Weiming Lu$^{1}$\thanks{\textsuperscript{$\ast$} Corresponding author}}\\
 $^{1}$College of Computer Science and Technology, Zhejiang University \\
 $^{2}$University of Science and Technology of China\\
  \texttt{\{syl, luwm\}@zju.edu.cn}
 }

\date{}

\begin{document}

\maketitle

\begin{abstract}

Named entity recognition (NER) is a well-studied task in natural language processing. Traditional NER research only deals with flat entities and ignores nested entities. The span-based methods treat entity recognition as a span classification task. Although these methods have the innate ability to handle nested NER, they suffer from high computational cost, ignorance of boundary information, under-utilization of the spans that partially match with entities, and difficulties in long entity recognition. To tackle these issues, we propose a two-stage entity identifier. First we generate span proposals by filtering and boundary regression on the seed spans to locate the entities, and then label the boundary-adjusted span proposals with the corresponding categories. Our method effectively utilizes the boundary information of entities and partially matched spans during training. Through boundary regression, entities of any length can be covered theoretically, which improves the ability to recognize long entities. In addition,  many low-quality seed spans are filtered out in the first stage, which reduces the time complexity of inference. Experiments on nested NER datasets demonstrate that our proposed method outperforms previous state-of-the-art models.

\end{abstract}

\section{Introduction}

\begin{figure}[h]
  \centering
  \includegraphics[width=\linewidth]{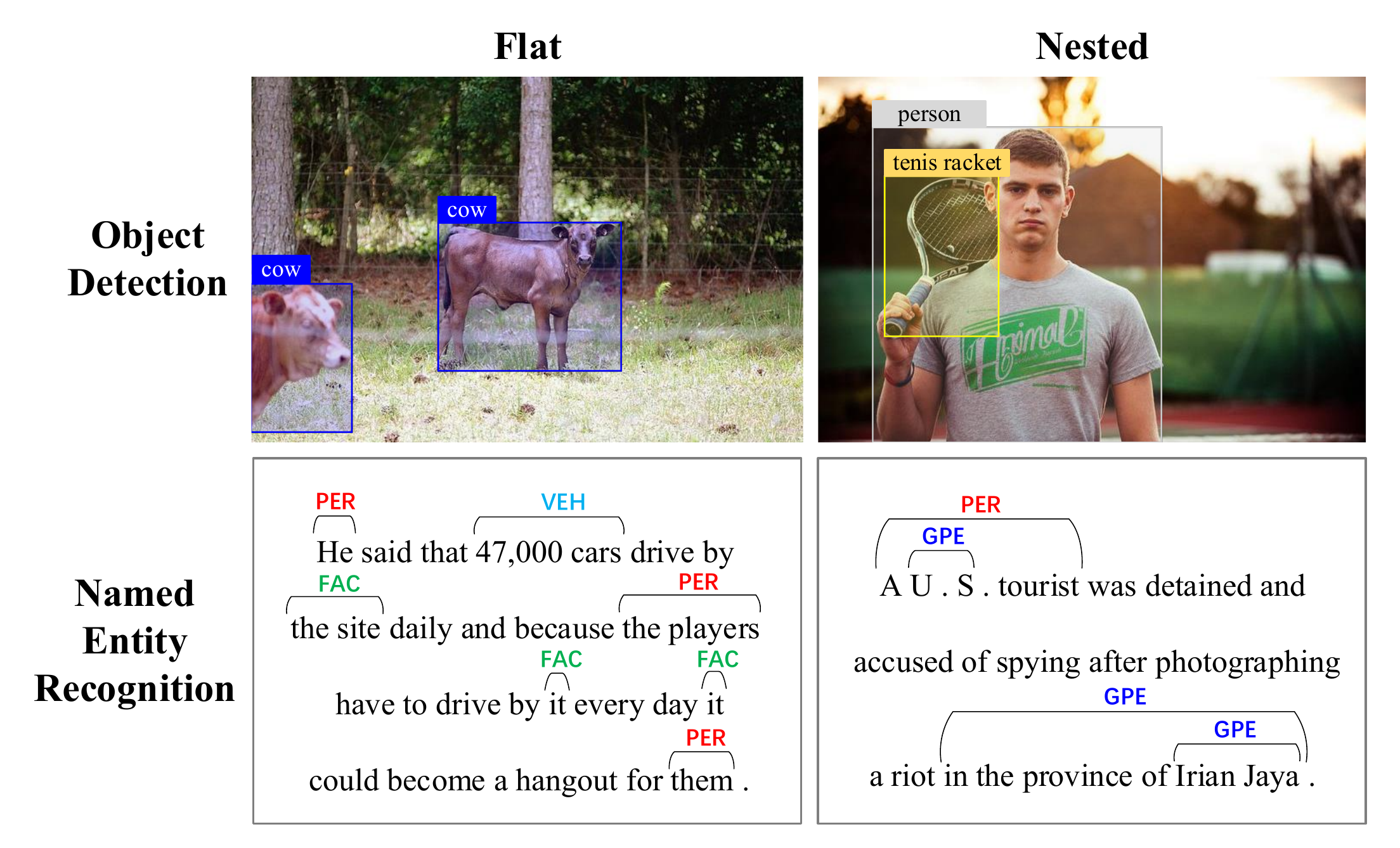}
  \caption{A Comparison of Named Entity Recognition and Object Detection. Examples of flat and nesetd entities or objects sampled from the COCO 2017 dataset and the ACE04 dataset, respectively.}
   \label{fig:example}
\end{figure}

Named entity recognition (NER) is a fundamental task in natural language processing, focusing on identifying the spans of text that refer to entities. NER is widely used in downstream tasks, such as entity linking \citep{ganea-hofmann-2017-deep, le-titov-2018-improving} and relation extraction \citep{li-ji-2014-incremental, miwa-bansal-2016-end}.

Previous works usually treat NER as a sequence labeling task, assigning a single tag to each token in a sentence. Such models lack the ability to identify nested named entities. Various approaches for nested NER have been proposed in recent years. Some works revised sequence models to support nested entities using different strategies \citep{alex-etal-2007-recognising, ju-etal-2018-neural, strakova-etal-2019-neural, wang-etal-2020-pyramid} and some works adopt the hyper-graph to capture all possible entity mentions in a sentence \citep{lu-roth-2015-joint, katiyar-cardie-2018-nested}. We focus on the span-based methods \citep{sohrab-miwa-2018-deep, zheng-etal-2019-boundary, Tan_Qiu_Chen_Wang_Huang_2020}, which treat named entity recognition as a classification task on a span with the innate ability to recognize nested named entities. For example, \citet{sohrab-miwa-2018-deep} exhausts all possible spans in a text sequence and then predicts their categories. However, these methods suffer from some serious weaknesses. First, due to numerous low-quality candidate spans, these methods require high computational costs. Then, it is hard to identify long entities because the length of the span enumerated during training is not infinite. Next, boundary information is not fully utilized, while it is important for the model to locate entities. Although some methods \citep{zheng-etal-2019-boundary, Tan_Qiu_Chen_Wang_Huang_2020} have used a sequence labeling model to predict boundaries, yet without dynamic adjustment, the boundary information is not fully utilized. Finally, the spans which partially match with entities are not effectively utilized. These methods simply treat the partially matched spans as negative examples, which can introduce noise into the model.

Different from the above studies, we observed that NER and object detection tasks in computer vision have a high degree of consistency. They both need to locate regions of interest (ROIs) in the context (image/text) and then assign corresponding categories to them. Furthermore, both flat NER and nested NER have corresponding structures in the object detection task, as shown in Figure \ref{fig:example}. 
For the flat structure, there is no overlap between entities or between objects. While for nested structures, fine-grained entities are nested inside coarse-grained entities, and small objects are nested inside large objects correspondingly. 
In computer vision, the two-stage object detectors \citep{10.1109/CVPR.2014.81,10.1109/ICCV.2015.169,10.1109/TPAMI.2016.2577031,10.5555/3157096.3157139,8237584,8578742} are the most popular object detection algorithm. They divide the detection task into two stages, first generating candidate regions, and then classifying and fine-tuning the positions of the candidate regions.

Inspired by these, we propose a two-stage entity identifier and treat NER as a joint task of boundary regression and span classification to address the weaknesses mentioned above.
In the first stage, we design a span proposal module, which contains two components: a filter and a regressor. The filter divides the seed spans into contextual spans and span proposals, and filters out the former to reduce the candidate spans. The regressor locates entities by adjusting the boundaries of span proposals to improve the quality of candidate spans. Then in the second stage, we use an entity classifier to label entity categories for the number-reduced and quality-improved span proposals. During training, to better utilize the spans that partially match with the entities, we construct soft examples by weighting the loss of the model based on the IoU. 
In addition, we apply the soft non-maximum suppression (Soft-NMS) \citep{8237855} algorithm to entity decoding for dropping the false positives.

Our main contributions are as follow:

\begin{itemize}
    \item Inspired by the two-stage detector popular in object detection, we propose a novel two-stage identifier for NER of locating entities first and labeling them later. We treat NER as a joint task of boundary regression and span classification.
    \item We make effective use of boundary information. Taking the identification of entity boundaries a step further, our model can adjust the boundaries to accurately locate entities. And when training the boundary regressor, in addition to the boundary-level SmoothL1 loss, we also use a span-level loss, which measures the overlap between two spans.
    \item During training, instead of simply treating the partially matched spans as negative examples, we construct soft examples based on the IoU. This not only alleviates the imbalance between positive and negative examples, but also effectively utilizes the spans which partially match with the ground-truth entities.
    \item Experiments show that our model achieves state-of-the-art performance consistently on the KBP17, ACE04 and ACE05 datasets, and outperforms several competing baseline models on F1-score by +3.08\% on KBP17, +0.71\% on ACE04 and +1.27\% on ACE05.
\end{itemize}

\section{Model}
\begin{figure*}[h]
  \centering
  \includegraphics[width=\linewidth]{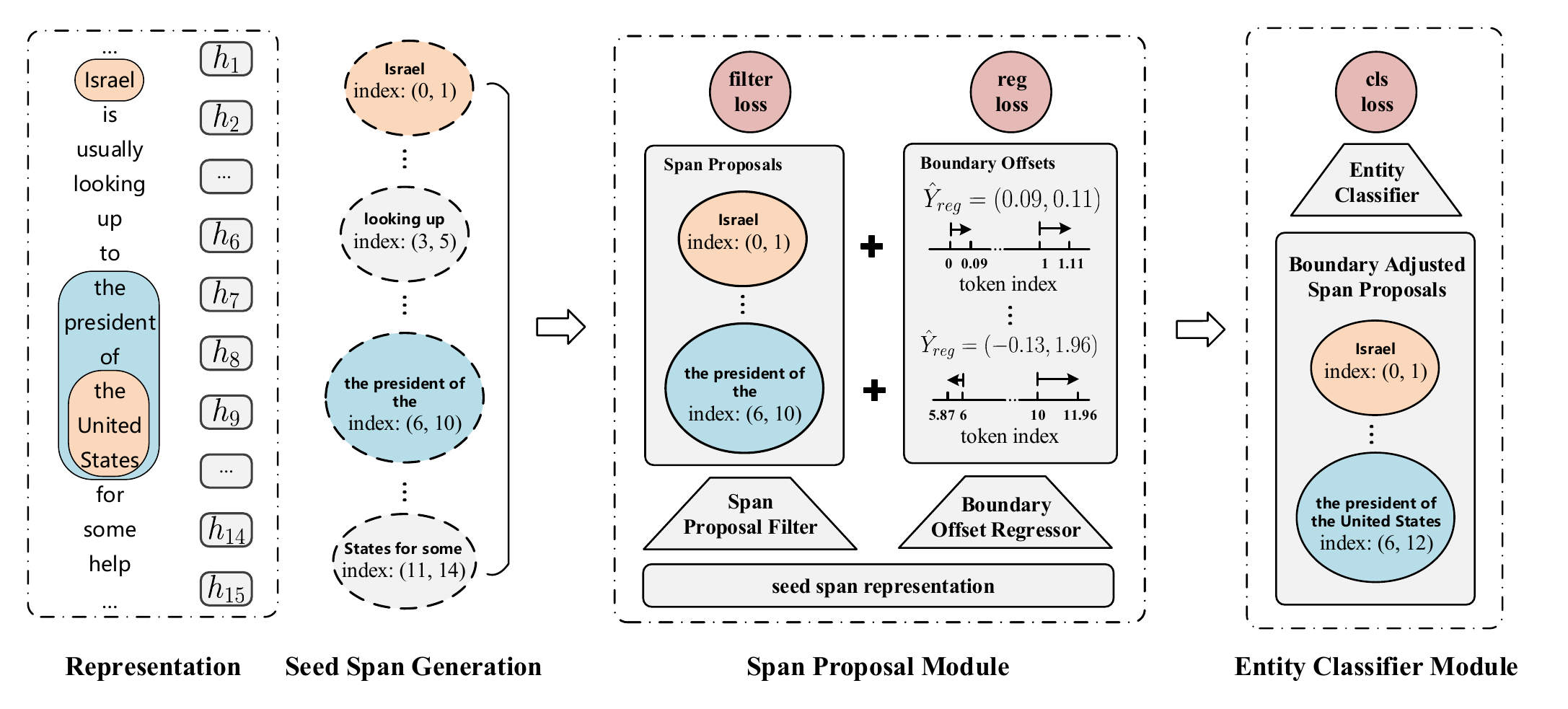}
  \caption{The overall architecture of the Two-stage Identifier. }
   \label{fig:overview}
\end{figure*}

Figure \ref{fig:overview} illustrates an overview of the model structure. We first obtain the word representation through the encoder and generate seed spans. Among these seed spans, some with higher overlap with the entities are the proposal spans, and others with lower overlap are the contextual spans. In the span proposal module, we use a filter to keep the proposal spans and drop the contextual spans. Meanwhile, a regressor regresses the boundary of each span to locate the left and right boundaries of entities. Next, we adjust the boundaries of the span proposals based on the output of the regressor, and then feed them into the entity classifier module. Finally, the entity decoder decodes the entities using the Soft-NMS algorithm. We will cover our model in the following sections.

\subsection{Token Representation}

Consider the $i$-th word in a sentence with $n$ words, we represent it by concatenating its word embedding $x^{w}_i$, contextualized word embedding $x^{lm}_i$, part-of-speech(POS) embedding $x^{pos}_i$ and character-level embedding $x^{char}_i$ together. The character-level embedding is generated by a BiLSTM module with the same setting as \citep{ju-etal-2018-neural}.
For the contextualized word embedding, we follow \citep{yu-etal-2020-named} to obtain the context-dependent embedding for a target token with one surrounding sentence on each side. Then, the concatenation of them is fed into another BiLSTM to obtain the hidden state as the final word representation $h_i\in \mathbb{R}^d$.

\subsection{Seed Span Generation}

Seed spans are subsequences sampled from a sequence of words. By filtering, adjusting boundaries, and classifying on them, we can extract entities from the sentence. Under the constraint of a pre-specified set of lengths, where the maximum does not exceed $L$, we enumerate all possible start and end positions to generate the seed spans. We denote the set of seed spans as $\mathcal{B}=\{b_0, \dots, b_K\}$, where $b_i = (st_i, ed_i)$ denotes $i$-th seed span, $K$ denotes the number of the generated seed spans, and $st_i$, $ed_i$ denote the start and end positions of the span respectively.

For training the filter and the regressor, we need to assign a corresponding category and regression target to each seed span. Specifically, we pair each seed span in $\mathcal{B}$ and the ground-truth entity with which the span has the largest IoU. The IoU measure the overlap between spans, defined as $ \operatorname{IoU}(A, B) =\frac{A\cap B}{A\cup B}$, where $A$ and $B$ are two spans. Then we divide them into positive and negative spans based on the IoU between the pair. The spans whose IoU with the paired ground truth is above the threshold $\alpha_1$ are classified as positive examples, and those less than threshold $\alpha_1$ are classified as negative examples. For the positive span, we assign it the same category $\hat{y}$ with the paired ground truth and compute the boundary offset $\hat{t}$ between them. For the negative span, we only assign a \textsc{None} label. We downsample the negative examples such that the ratio of positive to negative is 1:5.

\subsection{Span Proposal Module}

The quality of the generated seed spans is variable. If we directly input them into the entity classifier, it will lead to a lot of computational waste. High-quality spans have higher overlap with entities, while low-quality spans have lower overlap. We denote them as \textbf{span proposals} and \textbf{contextual spans}, respectively. Our Span Proposal module consists of two components: Span Proposal Filter and Boundary Regressor. The former is used to drop the contextual spans and keep the span proposals, while the latter is used to adjust the boundaries of the span proposals to locate entities.

\paragraph{Span Proposal Filter}

For the seed span $b_i(st_i, ed_i)$, we concatenate the maximum pooled span representation $h^p_i$ with the inner boundary word representations $(h_{st_i}, h_{ed_i})$ to obtain the span representation $h^{filter}_i$. 
Based on it we calculate the probability $p^{filter}_i$ that the span $b_i$ belongs to the span proposals, computed as follows:

\begin{equation}
    h^p_i = \operatorname{MaxPooling}(h_{st_i}, h_{st_i+1}, \dots, h_{ed_i})
\end{equation}

\begin{equation}
    h^{filter}_i = \left[h^p_i;h_{st_i}; h_{ed_i}\right]
\end{equation}

\begin{equation}
    p^{filter}_i = \operatorname{Sigmoid}\left(\operatorname{MLP}\left(h^{filter}_i\right)\right)
\end{equation}

\noindent where $[;]$ denotes the concatenate operation, $\operatorname{MLP}$ consists of two linear layers and a $\operatorname{GELU}$ \citep{DBLP:journals/corr/HendrycksG16} activation function.

\paragraph{Boundary Regressor}

Although the span proposal has a high overlap with the entity, it cannot hit the entity exactly. We design another boundary regression branch where a regressor locates entities by adjusting the left and right boundaries of the span proposals. The boundaries regression requires not only the information of span itself but also the outer boundary words. Thus we concatenate the maximum pooled span representation $h^p_i$ with the outer boundary word representations $(h_{st_i-1}, h_{ed_i+1})$ to obtain the span representation $h^{reg}_i$. Then we calculate the offsets $t_i$ of left and right boundaries:

\begin{equation}
    h^{reg}_i = \left[h^p_i;h_{st_i-1}; h_{ed_i+1}\right]
\end{equation}
\begin{equation}
    t_i =W_2\cdot \operatorname{GELU}(W_1h^{reg}_i+ b_1)+b_2
\end{equation}

\noindent where $W_1\in \mathbb{R}^{3d\times d}$, $W_2\in \mathbb{R}^{d\times2}$, $b_1\in \mathbb{R}^d$ and $b_2 \in \mathbb{R}^2$ are learnable parameters.

\subsection{Entity Classifier Module}

With the boundary offsets $t_i$ predicted by the boundary regressor, we adjust the boundaries of span proposals. The adjusted start postion $\widetilde{st}_i$ and end position $\widetilde{ed}_i$ of $b_i$ are calculated as follow:

\begin{equation}
    \widetilde{st}_i = \operatorname{max}(0, st_i + \left\lfloor t_i^{l}+\frac{1}{2}\right\rfloor)
\end{equation}
\begin{equation}
    \widetilde{ed}_i = \operatorname{min}(L-1, ed_i + \left\lfloor t_i^{r}+\frac{1}{2}\right\rfloor )
\end{equation}

\noindent where $t_i^{l}$ and $t_i^{r}$ denote the left and right offsets, respectively. As in the filter above, we concatenate the maximum pooled span representation $\widetilde{h}^p_i$ with the inner boundary word representations $(h_{\widetilde{st}_i}, h_{\widetilde{ed}_i})$. Then we perform entity classification:

\begin{equation}
    \widetilde{h}^p_i = \operatorname{MaxPooling}(h_{\widetilde{st}_i}, h_{\widetilde{st}_i+1}, \dots, h_{\widetilde{ed}_i})
\end{equation}
\begin{equation}
    h^{cls}_i = [\widetilde{h}^p_i;h_{\widetilde{st}_i}; h_{\widetilde{ed}_i}]
\end{equation}
\begin{equation}
    p_i = \operatorname{Softmax}\left(\operatorname{MLP}\left(h^{cls}_i )\right)\right)
\end{equation}

\noindent where $\operatorname{MLP}$ consists of two linear layers and a $\operatorname{GELU}$ activation function, as in the filter above.

For training the entity classifier, we need to reassign the categories based on the IoU between the new adjusted span proposal and paired ground-truth entity. Specifically, if the IoU between a span and its corresponding entity is higher than the threshold $\alpha_2$, we assign the span the same category with the entity, otherwise we assign it a \textsc{None} category and treat the span as a negative example. 

\subsection{Training Objective}

The spans that partially match with the entities are very important, but previous span-based approaches simply treat them as negative examples. Such practice not only fails to take advantage of these spans but also introduces noise into the model. We treat partially matched spans as soft examples by weighting its loss based on its IoU with the corresponding ground truth. For the $i$-th span $b_i$, the weight $w_i $ is calculated as follows:

\begin{equation}
\left\{
            \begin{array}{lr}
             \operatorname{IoU}(b_i,e_i)^\eta, &  \operatorname{IoU}(b_i,e_i)\geq \alpha \\
             
             \left(1-\operatorname{IoU}(b_i,e_i)\right)^\eta, & \operatorname{IoU}(b_i,e_i)<\alpha\\
             \end{array}
\right.
\label{eq:weight}
\end{equation}

\noindent where $\alpha \in \{\alpha_1, \alpha_2\}$ denotes the IoU threshold used in the first or the second stage and $e_i$ denotes corresponding ground-truth entity of $b_i$. $\eta$ is a focusing parameter that can smoothly adjust the rate at which partially matched examples are down-weighted. We can find that if we set $\eta=0$, the above formula degenerates to a hard one. Also, if a span does not overlap with any entity or match exactly with some entity, the loss weight $w_i=1$.


Then, we calculate the losses for the span proposal filter, boundary regressor and entity classifier respectively. For the span proposal filter, we use focal loss \citep{8237586} to solve the imbalance problem:

\begin{equation}
    \begin{aligned}
    \mathcal{L}_{filter} &= -\sum_i w_i \mathbb{I}_{\hat{y}\neq0}(1-p_i^{filter})^\gamma\log(p_i^{filter})\\ &+w_i\mathbb{I}_{\hat{y}= 0}(p_i^{filter})^\gamma  \log(1-p_i^{filter})
    \end{aligned}
\end{equation}

\noindent where $w_i$ is the weight of $i$-th example calculated at Equation \ref{eq:weight} and $\gamma$ denotes focusing parameter of focal loss.  For the boundary regressor, the loss consists of two components, the smooth L1 loss at the boundary level and the overlap loss at the span level, calculated as follows:

\begin{equation}
\mathcal{L}_{reg}\left(\hat{t}, t\right)=\mathcal{L}_{f1}+\mathcal{L}_{olp}
\end{equation}

\begin{equation}
\mathcal{L}_{f1}\left(\hat{t}, t\right)=\sum_i \sum_{j \in\{l, r\}} \operatorname{ smoothL1}\left(\hat{t}^j_{i}, t^j_{i}\right)
\end{equation}




\begin{equation}
\mathcal{L}_{olp}=\sum_i \left(1- \frac{\min\left(d_i\right) - \max\left(e_i\right) }{\max \left(d_i\right) - \min\left(e_i\right)}\right)
\end{equation}

\noindent where $d_i =  \left\{\widetilde{ed}_i,\hat{ed}_i\right\}$, $e_i=\left\{\widetilde{st}_i, \hat{st}_i\right\}$. $\hat{st}_i$, $\hat{ed}_i$, $\hat{t}_i^{l}$ and $\hat{t}_i^{r}$ denote the ground-truth left boundary, right boundary, left offset and right offset, respectively. For the entity classifier, we simply use the cross-entropy loss:



\begin{equation}
    \begin{aligned}
    \mathcal{L}_{cls} = \sum_i w_i\operatorname{CELoss}(\hat{y}, p_i)
    \end{aligned}
\end{equation}

\noindent where $w_i$ is the weight of $i$-th example calculated at Equation \ref{eq:weight}. We train the filter, regressor and classifier jointly, thus the total loss is computed as:
\begin{equation}
\mathcal{L} = \lambda_1 \mathcal{L}_{filter}+ \lambda_2 \mathcal{L}_{reg} +\lambda_3 \mathcal{L}_{cls}
\end{equation}
where $\lambda_1$, $\lambda_2$ and $\lambda_3$ are the weights of filter, regressor and classifier losses respectively.

\subsection{Entity Decoding}

In the model prediction phase, after the above steps, we get the classification probability and boundary offset regression results for each span proposal. Based on them, we need to extract all entities in the sentence (i.e., find the exact start and end positions of the entities as well as their corresponding categories). We assign label $y_i = \operatorname{argmax}(p_i)$ to span $s_i$ and use $score_i = \operatorname{max}(p_i)$ as the confidence of span $s_i$ belonging to the $y_i$ category.

Now for each span proposal, our model has predicted the exact start and end positions, the entity class and the corresponding score, denoted as $s_i=(l_i,r_i,y_i,score_i)$. Given the score threshold $\delta$ and the set of span proposals  $\mathcal{S} = \{s_1,\dots,s_N\}$, where $N$ denotes as the number of span proposals, we use the Soft-NMS \citep{8237855} algorithm to filter the false positives. As shown in Algorithm \ref{alg:sof-nms}, we traverse the span proposals by the order of their score 
(the traversal term is denoted as $s_i$) and then adjust the scores of other span proposals $s_j$ to $f(s_i,s_j)$, which is defined as:

\begin{equation}
\left\{
            \begin{array}{lr}
             score_j * u, &  \operatorname{IoU}(s_i,s_j)\geq k \\
             score_j, & \operatorname{IoU}(s_i,s_j)<k\\
             \end{array}
\right.
\end{equation}

\noindent where $u\in(0, 1)$ denotes the decay coefficient of the score and $k$ denotes is the IoU threshold. Then we keep all span proposals with a $score > \delta$ as the final extracted entities.

\begin{algorithm}[]
\KwIn{$\mathcal{S} = \{s_i,\dots,s_N\}$, $\delta$, where $s_i=(l_i,r_i,y_i,score_i)$}
\KwOut{$\mathcal{O}$}
 $\mathcal{O} \leftarrow \{\}$\;
 $\operatorname{Sort}( \mathcal{S} )$ by the $score$ of each element in descend order\;
 \For{$s_i$ in $\mathcal{S}$}{
$\mathcal{O} \leftarrow \mathcal{O} \cup \{ s_i \}$ \;
\For{$s_j$ in $\mathcal{S}\left[i:N\right]$}{
$\mathcal{S} \leftarrow \mathcal{S} - \{ s_{j} \}$\;
$s_j \leftarrow  (l_j,r_j,y_j, f(s_i, s_j))$\;
$\operatorname{Insert}\left(\mathcal{S},k, s_j\right)$ where $k$ denotes the insertion position of $s_j$ in $\mathcal{S}$ ordered by $score$\;}
}

 \caption{Soft-NMS Algorithm}
 \label{alg:sof-nms}
\end{algorithm}

\section{Experiment Settings}
\subsection{Datasets}

To provide empirical evidence for effectiveness of the proposed model, we conduct our experiments on four nested NER datasets: ACE04 \footnote{{\ https://catalog.ldc.upenn.edu/LDC2005T09}}, ACE05 \footnote{{\ https://catalog.ldc.upenn.edu/LDC2006T06}}, KBP17\footnote{{\ https://catalog.ldc.upenn.edu/LDC2019T02}} and GENIA \footnote{{\ http://www.geniaproject.org/genia-corpus}}. Please refer to Appendix \ref{app:statistic} for statistical information  about the datasets.

\paragraph{ACE 2004 and ACE 2005} \citep{doddington-etal-2004-automatic, 2005-automatic} are two nested datasets, each of them contains 7 entity categories. We follow the same setup as previous work \citet{katiyar-cardie-2018-nested, lin-etal-2019-sequence} split them into train, dev and test sets by 8:1:1.

\paragraph{KBP17} \citep{DBLP:conf/tac/JiPZNMMC17} has 5 entity categories, including GPE, ORG, PER, LOC, and FAC. We follow \citet{lin-etal-2019-sequence} to split all documents into 866/20/167 documents for train/dev/test set.

\paragraph{GENIA} \citep{10.5555/1289189.1289260} is a biology nested named entity dataset and contains five entity types, including DNA, RNA, protein, cell line, and cell type categories. Following \citet{yu-etal-2020-named}, we use 90\%/10\% train/test split.

\subsection{Evaluation Metrics}

We use strict evaluation metrics that an entity is confirmed correct when the entity boundary and the entity label are correct simultaneously. We employ precision, recall and F1-score to evaluate the performance.

\subsection{Parameter Settings}
In most experiments, we use GloVE \citep{pennington-etal-2014-glove} and BERT \citep{devlin-etal-2019-bert} in our encoder. For the GENIA dataset, we replace GloVE with BioWordvec \citep{chiu-etal-2016-train}, BERT with BioBERT \citep{10.1093/bioinformatics/btz682}. The dimensions for $x^{w}_i$, $x^{lm}_i$, $x^{pos}_i$, $x^{char}_i$ and $h_i$ are 100, 1024, 50, 50 and 1024, respectively. For all datasets, we train our model for 35 epochs and use the Adam Optimizer with a linear warmup-decay learning rate schedule, a dropout before the filter, regressor and entity classifier with a rate of 0.5. See Appendix \ref{app:nestedner} for more detailed parameter settings and baseline models we compared \footnote{\ Our code is available at \url{https://github.com/tricktreat/locate-and-label}.}.

\section{Results and Comparisons}
\subsection{Overall Evaluation}

Table \ref{tab:nested} illustrates the performance of the proposed model as well as baselines on ACE04, ACE05, GENIA and KBP17. Our model outperforms the state-of-the-art models consistently on three nested NER datasets. Specifically, the F1-scores of our model advance previous models by +3.08\%, +0.71\%, +1.27\% on KBP17, ACE04 and ACE05 respectively. And on GENIA, we achieve comparable performance. We analyze the performance on entities of different lengths on ACE04, as shown in Table \ref{tab:length}. 
We observe that the model works well on the entities whose lengths are not enumerated during training.
For example, although entities of length 6 are not enumerated, while those of length 5 and 7 are enumerated, our model can achieve a comparable F1-score for entities of length 6. 
In particular, the entities whose lengths exceed the maximum length (15) enumerated during training, are still well recognized. 
This verifies that our model has the ability to identify length-uncovered entities and long entities by boundary regression. We also evaluated our model on two flat NER datasets, as shown in Appendix \ref{app:flatner}.

\begin{table}[t]
\centering
\small
\begin{tabular}{lccc}
\toprule
\multirow{2}{*}{Model}   & \multicolumn{3}{c}{ACE04}  \\
 \cmidrule(lr){2-4} 
& Pr.  & Rec. & F1  \\
\midrule

\citet{katiyar-cardie-2018-nested} 
&  73.60 &  71.80 &  72.70 \\
\citet{shibuya-hovy-2020-nested}   & 83.73 & 81.91 & 82.81\\
\citet{strakova-etal-2019-neural}  & - & - &  84.40  \\
\citet{wang-etal-2020-pyramid}       & 86.08  & 86.48  & 86.28     \\
\citet{yu-etal-2020-named}       & 87.30  & 86.00  & 86.70     \\
\midrule
Ours    & \textbf{87.44}  & \textbf{87.38}  & \textbf{87.41}  \\
\bottomrule
\toprule
\multirow{2}{*}{Model}   & \multicolumn{3}{c}{ACE05}  \\
 \cmidrule(lr){2-4} 
& Pr.  & Rec. & F1  \\
\midrule



\citet{katiyar-cardie-2018-nested}  & 70.60 & 70.40 & 70.50 \\
\citet{lin-etal-2019-sequence}       & 76.20  & 73.60 & 74.90  \\
\citet{luo-zhao-2020-bipartite}       & 75.00  & 75.20  & 75.10    \\
\citet{strakova-etal-2019-neural}  & - & - &  84.33  \\
\citet{wang-etal-2020-pyramid}       & 83.95  & 85.39  & 84.66     \\
\citet{yu-etal-2020-named}       & 85.20  & 85.60  & 85.40     \\
\midrule
Ours    & \textbf{86.09}   & \textbf{87.27} & \textbf{86.67}  \\

\bottomrule
\toprule
\multirow{2}{*}{Model}   & \multicolumn{3}{c}{KBP17}  \\
 \cmidrule(lr){2-4} 
& Pr.  & Rec. & F1  \\
\midrule

\citet{DBLP:conf/tac/JiPZNMMC17}       & 76.20 & 73.00 & 72.80  \\
\citet{lin-etal-2019-sequence}       & 77.70  & 71.80 & 74.60  \\
\citet{luo-zhao-2020-bipartite}       & 77.10 & 74.30 & 75.60  \\
\citet{li-etal-2020-unified} & 80.97  & 81.12  & 80.97  \\
\midrule
Ours    & \textbf{85.46}  & \textbf{82.67}  & \textbf{84.05}  \\
\bottomrule
\toprule
\multirow{2}{*}{Model}   & \multicolumn{3}{c}{GENIA}  \\
 \cmidrule(lr){2-4} 
& Pr.  & Rec. & F1  \\
\midrule




\citet{lin-etal-2019-sequence}       & 75.80  & 73.90 & 74.80  \\
\citet{luo-zhao-2020-bipartite}       & 77.40  & 74.60  & 76.00    \\
\citet{wang-etal-2020-hit}       & 78.10  & 74.40  & 76.20  \\
\citet{strakova-etal-2019-neural}  & - & - &  78.31 \\
\citet{wang-etal-2020-pyramid}       & 79.45  & 78.94  & 79.19     \\
\citet{yu-etal-2020-named}       & 81.80  & 79.30  & 80.50     \\
\midrule
Ours    & 80.19 & \textbf{80.89}  & \textbf{80.54}  \\
\bottomrule

\end{tabular}
\caption{Results for \textit{\textbf{nested}} NER tasks}
\label{tab:nested}
\end{table}

\begin{table}[!]
\centering
\small
\begin{tabular}{c>{\centering\arraybackslash}m{1.0cm}>{\centering\arraybackslash}m{1.0cm}>{\centering\arraybackslash}m{1.0cm}c}
\toprule
\multicolumn{1}{c}{\multirow{2}{*}{Length} }  & \multicolumn{4}{c}{ACE04}  \\
 \cmidrule(lr){2-5} 
& Pr.  & Rec. & F1 & Support  \\
\midrule
 1 &   89.62 &   90.98 &   90.30 & 1519 \\
 2 &   87.93 &   86.10 &   87.01 & 626 \\
 3 &   89.67 &   84.59 &   87.06 & 318 \\
 4 &   79.04 &   88.59 &   83.54 & 149 \\
 5 &   85.58 &   83.18 &   84.36 & 107 \\
  \textbf{6} &   \textbf{84.62} &   \textbf{86.84} &   \textbf{85.71} & \textbf{76}\\
 7 &   85.07 &   85.07 &   85.07 & 67 \\
  \textbf{8} &    \textbf{79.31} &    \textbf{79.31} &    \textbf{79.31} &  \textbf{29} \\
9 &   81.48 &   73.33 &   77.19 & 30 \\
 \textbf{10} &    \textbf{76.47} &    \textbf{76.47} &    \textbf{76.47} &  \textbf{17} \\
 11 &   68.75 &   68.75 &   68.75 & 16 \\
  \textbf{12} &    \textbf{66.67} &    \textbf{80.00} &    \textbf{72.73} &  \textbf{15} \\
 13 &  100.00 &   85.71 &   92.31 & 7 \\
  \textbf{14} &    \textbf{55.56} &    \textbf{83.33} &    \textbf{66.67} &  \textbf{6} \\
 15 &   55.56 &   71.43 &   62.50 & 7 \\
  \textbf{16} &    \textbf{80.00} &    \textbf{57.14} &   \textbf{66.67} & \textbf{7} \\
  \textbf{17} &    \textbf{66.67} &    \textbf{80.00} &    \textbf{72.73} &  \textbf{5} \\
 \textbf{18} &    \textbf{83.33} &    \textbf{83.33} &    \textbf{83.33} &  \textbf{6} \\
 \textbf{19} &    \textbf{33.33} &    \textbf{33.33} &    \textbf{33.33} &  \textbf{3} \\
 $\mathbf{\geq20}$ &    \textbf{66.67} &    \textbf{24.00} &    \textbf{35.29} &  \textbf{25} \\
\midrule
\multicolumn{1}{c}{All} &   87.46 &   87.35 &   87.41 & 3035 \\
\bottomrule
\end{tabular}
\caption{A comparison of recognition F1-score on entities of different lengths. Regular rows indicate that the entity lengths are enumerated, while bold ones indicate that the entity lengths are not enumerated.}
\label{tab:length}
\end{table}

\subsection{Ablation Study}

We choose the ACE04 and KBP17 datasets to conduct several ablation experiments to elucidate the main components of our proposed model. To illustrate the performance of the model on entities of different lengths, we divide the entities into three groups according to their lengths. The results are shown in Table \ref{tab:ablation}. Firstly, we observe that the boundary regressor is very effective for the identification of long entities. Lack of the boundary regressor leads to a decrease in F1-score for long entities ($L\geq10$) on ACE04 by 36.73\% and KBP17 by 30.54\%. Then, compared with the \textit{w/o filter} setting, the F1-scores of our full model on the two datasets improved by 0.52\% and 0.75\%, respectively. In addition, experimental results also demonstrate that the soft examples we constructed are effective. This allows the model to take full advantage of the information of partially matched spans in training, improving the F1-score by 0.87\% on ACE04 and 0.16\% on KBP17. However, Soft-NMS play a limited role and improve the model performance only a little. We believe that text is sparse data compared to images and the number of false positives predicted by our model is quite small, so the Soft-NMS can hardly perform the role of a filter.

\begin{table*}[h]
\centering
\small
\begin{tabular}{lccccccccc}
\toprule
\multirow{2}{*}{Model}   & \multicolumn{4}{c}{F1-score on ACE04}  & \multicolumn{4}{c}{F1-score on KBP17}  \\
 \cmidrule(lr){2-5}  \cmidrule(lr){6-9}  
& 1$\leq$L$<$5  & 5$\leq$L$<$10 & L$\geq$10 & ALL & 1$\leq$L$<$5  & 5$\leq$L$<$10 & L$\geq$10 & ALL  \\
\midrule
support & 2612 & 309 & 114 & 3035 & 11594 & 756 & 250 & 12600\\
\midrule
Full model   & \textbf{88.73} & 83.71 & \textbf{66.06} & \textbf{87.41} & \textbf{85.52} & 67.67 & 58.58 & \textbf{84.05}\\
\quad w/o regressor  & 88.63 & 66.41 & 29.33 & 85.18
& 83.99 & 50.50 & 28.04 & 82.54 \\

\quad w/o filter & 88.35 & 83.87 & 60.55 & 86.89
&   84.77 & 67.04 & 59.06 & 83.30\\ 

\quad w/o filter \& regressor & 88.59 & 65.65 & 31.08 & 85.12 
& 85.28 &51.76 &26.03 & 82.85 \\ 
\quad w/o soft-NMS & 88.66 & 83.50 & 65.16 & 87.28 & 85.49 & 67.62 & 58.77 & 84.02\\
\quad w/o soft examples  & 88.39 & 80.39 & 55.96 & 86.54
& 85.27 & 68.95 &60.85 &83.89\\ 
\bottomrule
\end{tabular}
\caption{Ablation study on ACE04 and KBP17. To compare the performance of the model on entities of different lengths, we divided the entities into three groups: $1\leq L<5$, $5\leq L<10$ and $L \geq 10$.}
\label{tab:ablation}
\end{table*}

\subsection{Time Complexity}

Theoretically, the number of possible spans of a sentence of length $N$ is $\frac{N(N+1)}{2}$. Previous span-based methods need to classify almost all spans into corresponding categories, which leads to the high computational cost with $O(cN^2)$ time complexity where $c$ is the number of categories. 
The words in a sentence can be divided into two categories: contextual words and entity words.
Traditional approaches waste a lot of computation on the spans composed of contextual words. However, our approach retains only the span proposals containing entity words by the filter, and the time complexity is $O(N^2)$. Although in the worst case the model keeps all seed spans, generating $\frac{N(N+1)}{2}$ span proposals, we observe that we generate approximately three times as many span proposals as the entities in practice. Assuming that the number of entities in the sentence is $k$, the total time complexity of our model is $O(N^2 + ck)$ where $k << N^2$.

\begin{table*}
\small
\centering
  \begin{tabular}{m{15.5cm}}
  \toprule
    \textcolor{red}{[\textsuperscript{3}}\textcolor{blue}{[\textsuperscript{3}}\textcolor{red}{[\textsuperscript{2}}\textcolor{blue}{[\textsuperscript{2}}\textcolor{red}{[\textsuperscript{1}}\textcolor{blue}{[\textsuperscript{1}}united nations\textcolor{blue}{$^1$]$_{\text{ORG}}$}\textcolor{red}{$^1$]$_{\text{ORG}}$} secretary general\textcolor{blue}{$^2$]$_{\text{PER}}$}\textcolor{red}{$^2$]$_{\text{PER}}$} kofi annan\textcolor{blue}{$^3$]$_{\text{PER}}$}\textcolor{red}{$^3$]$_{\text{PER}}$} today discussed plans for the summit with \textcolor{red}{[\textsuperscript{1}}\textcolor{blue}{[\textsuperscript{1}}the host\textcolor{blue}{$^1$]$_{\text{PER}}$}\textcolor{red}{$^1$]$_{\text{PER}}$} , \textcolor{red}{[\textsuperscript{3}}\textcolor{blue}{[\textsuperscript{3}}\textcolor{red}{[\textsuperscript{2}}\textcolor{blue}{[\textsuperscript{2}}\textcolor{red}{[\textsuperscript{1}}\textcolor{blue}{[\textsuperscript{1}}egyptian\textcolor{blue}{\textsuperscript{1}]$_{\text{GPE}}$}\textcolor{red}{\textsuperscript{1}]$_{\text{GPE}}$} president\textcolor{blue}{\textsuperscript{2}]$_{\text{PER}}$}\textcolor{red}{\textsuperscript{2}]$_{\text{PER}}$} hosni mubarak\textcolor{blue}{\textsuperscript{3}]$_{\text{PER}}$}\textcolor{red}{\textsuperscript{3}]$_{\text{PER}}$} .\\
\midrule
{\color{red}[\textsuperscript{1}}{\color{blue}[\textsuperscript{1}}Separatists{\color{blue}\textsuperscript{1}]$_{\text{PER}}$}{\color{red}\textsuperscript{1}]$_{\text{PER}}$} have fought since 1975 for independence in {\color{red}[\textsuperscript{3}}{\color{blue}[\textsuperscript{3}}Aceh , {\color{red}[\textsuperscript{1}}{\color{blue}[\textsuperscript{1}}which{\color{blue}\textsuperscript{1}]$_{\text{GEP}}$}{\color{red}\textsuperscript{1}]$_{\text{GEP}}$} is rich in oil and gas and has {\color{red}[\textsuperscript{2}}{\color{blue}[\textsuperscript{2}}a population of {\color{blue}[\textsuperscript{1}}about 4 . 1 million people{\color{blue}\textsuperscript{1}]$_{\text{PER}}$}{\color{blue}\textsuperscript{2}]$_{\text{PER}}$}{\color{red}\textsuperscript{2}]$_{\text{PER}}$}{\color{blue}\textsuperscript{3}]$_{\text{GEP}}$}{\color{red}\textsuperscript{3}]$_{\text{GEP}}$} .\\

    \midrule
    {\color{red}[\textsuperscript{2}}{\color{blue}[\textsuperscript{2}}The {\color{blue}[\textsuperscript{1}}US{\color{blue}\textsuperscript{1}]$_{\text{GPE}}$} Supreme Court{\color{blue}\textsuperscript{2}]$_{\text{ORG}}$}{\color{red}\textsuperscript{2}]$_{\text{ORG}}$} will hear arguments from {\color{red}[\textsuperscript{1}}{\color{blue}[\textsuperscript{1}}both sides{\color{blue}\textsuperscript{1}]$_{\text{PER}}$}{\color{red}\textsuperscript{1}]$_{\text{ORG}}$} on Friday and {\color{red}[\textsuperscript{2}}{\color{blue}[\textsuperscript{2}}{\color{red}[\textsuperscript{1}}{\color{blue}[\textsuperscript{1}}Florida{\color{blue}\textsuperscript{1}]$_{\text{GPE}}$}{\color{red}\textsuperscript{1}]$_{\text{GPE}}$} ' s    {\color{red}[\textsuperscript{1}}{\color{blue}[\textsuperscript{1}}Leon County{\color{blue}\textsuperscript{1}]$_{\text{GPE}}$}{\color{red}\textsuperscript{1}]$_{\text{GPE}}$} Circuit Court{\color{blue}\textsuperscript{2}]$_{\text{ORG}}$}{\color{red}\textsuperscript{2}]$_{\text{ORG}}$} will consider the arguments on disputed    {\color{red}[\textsuperscript{1}}{\color{blue}[\textsuperscript{1}}state{\color{blue}\textsuperscript{1}]$_{\text{GPE}}$}{\color{red}\textsuperscript{1}]$_{\text{GPE}}$} ballots on Saturday .\\
  \bottomrule
 \end{tabular}
 \caption{Cases Study. Blue brackets indicate entities predicted by the model, red brackets indicate true entities, the labels in the lower right corner indicate the type of entity, and the superscripts indicate the level of the nesting.}
  \label{tab:error}
\end{table*}

\section{Case Study}

Examples of model predictions are shown in Table \ref{tab:error}. The first line illustrates that our model can recognize entities with multi-level nested structures. We can see that the three nested entities from inside to outside are {\textit{united nations secretary general kofi annan}}, {\textit{united nations secretary general}} and {\textit{united nations}}, all of which can be accurately recognized by our model. The second line illustrates that our model can recognize long entities well, although trained without seed spans of the same length as it. The long entity {\textit{Aceh, which is rich in oil and gas and has a  population of about 4.1 million people}}, with a length of 20, exceeds the maximum length of generated seed spans, but can still be correctly located and classified.
However, our model has difficulties in resolving ambiguous entity references. As shown in the third line, our model incorrectly classifies the reference phrase \textit{both sides}, which refers to \textsc{Org}, into the \textsc{Per} category.

\section{Related Work}

\subsection{Nested Named Entity Recognition}
NER is usually modeled as a sequence labeling task, and a sequence model (e.g., LSTM-CRF \citep{huang2015bidirectional}) is employed to output the sequence of labels with maximum probability. However, traditional sequence labeling models cannot handle nested structures because they can only assign one label to each token. In recent years, several approaches have been proposed to solve the nested named entity recognition task, mainly including tagging-based \citep{alex-etal-2007-recognising, wang-etal-2020-pyramid}, hypergraph-based \citep{muis-lu-2017-labeling, katiyar-cardie-2018-nested}, and span-based \citep{sohrab-miwa-2018-deep, zheng-etal-2019-boundary} approaches. The tagging based nested NER model transforms the nested NER task into a special sequential tagging task by designing a suitable tagging schema. Layered-CRF \citep{alex-etal-2007-recognising} dynamically stacks flat NER layers to identify entities from inner to outer. Pyramid \citep{wang-etal-2020-pyramid} designs a pyramid structured tagging framework that uses CNN networks to identify entities from the bottom up. The hypergraph-based model constructs the hypergraph by the structure of nested NER and decodes the nested entities on the hypergraph. \citet{lu-roth-2015-joint} is the first to propose the use of Mention Hypergraphs to solve the overlapping mentions recognition problem.
\citet{katiyar-cardie-2018-nested} proposed hypergraph representation for the nested NER task and learned the hypergraph structure in a greedy way by LSTM networks.
The span-based nested NER model first extracts the subsequences (spans) in a sequence and then classifies these spans. Exhaustive Model \citep{sohrab-miwa-2018-deep} exhausts all possible spans in a text sequence and then predicts their classes. \citet{zheng-etal-2019-boundary, Tan_Qiu_Chen_Wang_Huang_2020} took a sequence labeling model to identify entity boundaries and then predicted the categories of boundary-relevant regions. 
Different from the above methods,
some works adopt the methods from other tasks. For example, \citet{yu-etal-2020-named} 
reformulated NER as a structured prediction task and adopted a biaffine model for nested and flat NER. While \citet{li-etal-2020-unified} treated NER as a reading comprehension task, and constructed type-specific queries to extract entities from the context.

\subsection{Object Detection}

Object detection is a computer vision technique that can localize and identify objects in an image. With this identification and localization, object detection can determine the exact location of objects while assigning them categories. Neural-based object detection algorithms are divided into two main categories: one-stage and two-stage approach. The one-stage object detector densely proposes anchor boxes by covering the possible positions, scales, and aspect ratios, and then predicts the categories and accurate positions based on them in a single-shot way, such as OverFeat \citep{sermanet2013overfeat}, YOLO \citep{7780460} and SSD \citep{liu2016ssd}. The two-stage object detector can be seen as an extension of the dense detector and has been the most dominant object detection algorithm for many years \citep{10.1109/CVPR.2014.81,10.1109/ICCV.2015.169,10.1109/TPAMI.2016.2577031,10.5555/3157096.3157139,8237584,8578742}. It first obtains sparse proposal boxes containing objects from a dense set of region candidates, and then adjusts the position and predicts a category for each proposal. 

\section{Conclusion}
In this paper, we treat NER as a joint task of boundary regression and span classification and propose a two-stage entity identifier.
First we generate span proposals through a filter and regressor, then classify them into the corresponding categories. Our proposed model can make full use of the boundary information of entities and reduce the computational cost. Moreover, by constructing soft samples during training, our model can exploit the spans that partially match with the entities. Experiments illustrate that our method achieves state-of-the-art performance on several nested NER datasets. For future work, we will combine named entity recognition and object detection tasks, and try to use a unified framework to address joint identification on multimodal data.

\section*{Acknowledgments}

This work is supported by the Key Research and Development Program of Zhejiang Province, China(No. 2021C01013), the National Key Research and Development Project of China (No. 2018AAA0101900), the Chinese Knowledge Center of Engineering Science and Technology (CKCEST) and MOE Engineering Research Center of Digital Library.

\bibliographystyle{acl_natbib}
\bibliography{acl2021}

\clearpage
\newpage

\appendix

\section{Experiments on Nested NER}

\label{app:nestedner}
\subsection{Statistics of Nested Datasets}
\label{app:statistic}
In Table \ref{tab:statistics}, We report the number of sentences, the number of sentences containing nested entities, the average sentence length, the total number of entities, the number of nested entities and the nesting ratio on the ACE04, ACE05, GENIA and KBP17 datasets.

\begin{table*}[!t]
\centering
\small
\begin{tabular}{lccccccccccc}
\toprule
\multirow{2}{*}{Dataset Statistics}   & \multicolumn{3}{c}{ACE04}& \multicolumn{3}{c}{ACE05} & \multicolumn{3}{c}{KBP17} & \multicolumn{2}{c}{GENIA}  \\
 \cmidrule(lr){2-4}  \cmidrule(lr){5-7} \cmidrule(lr){8-10} \cmidrule(lr){11-12}  
& Train  & Dev & Test & Train  & Dev & Test & Train  & Dev & Test & Train   & Test  \\
\midrule
\# sentences &  6200 &  745 &  812 &  7194 &  969 &  1047 &  10546 &  545 & 4267 &  16692 &   1854 \\
\# sent. nested entities  &  2712 &  294 &  388 &  2691 &  338 &  320 &  2809 &  182 &  1223 &  3522 &   446 \\
avg sentence length &  22.50 &  23.02 &  23.05 &  19.21 &  18.93 &  17.2 &  19.62 &  20.61 &  19.26 &  25.35 &    25.99 \\
\# total entities &  22204 &  2514 &  3035 &  24441 &  3200 &  2993 &  31236 &  1879 &  12601 &  50509 &    5506 \\
\# nested entities &  10149 &  1092 & 1417  & 9389 &  1112 &  1118 &  8773 &  605 &  3707 &  9064 &    1199 \\
nested percentage (\%) &  45.71 & 46.69 &  45.61 & 38.41 & 34.75 &  37.35 &  28.09 & 32.20 &  29.42 &  17.95 &    21.78 \\
\bottomrule

\end{tabular}
\caption{Statistics of the datasets used in the experiments.}
\label{tab:statistics}
\end{table*}

\subsection{Baseline Methods}
We use the following models as baselines for nested NER:

\begin{itemize}
    \item \textbf{Biaffine} \citep{yu-etal-2020-named} reformulates NER as a structured prediction task and adopts a dependency parsing approach for NER.
    \item \textbf{Pyramid} \citep{wang-etal-2020-pyramid} consists of a stack of inter-connected layers. Each layer predicts whether a text region of certain length is a complete entity mention.
    \item \textbf{BiFlaG} \citep{yu-etal-2020-named} designs a bipartite flat-graph network with two interacting subgraph modules for outermost entities and inner entities, respectively.
    \item \textbf{HIT} \citep{wang-etal-2020-hit} leverages the head-tail pair and token interaction to express the nested entities.
    \item \textbf{ARN} \citep{lin-etal-2019-sequence} designs a sequence-to-nuggets architecture by modeling and levraging the head-driven phrase structures of entity mentions.
    \item \textbf{Seq2seq} \citep{strakova-etal-2019-neural} views the nested NER as a sequence-to-sequence problem.
    \item \textbf{KBP17-Best} \citep{DBLP:conf/tac/JiPZNMMC17} gives an overview of the Entity Discovery task and reports previous best results for the task of nested NER.
\end{itemize}

We didn't compare our model with BERT-MRC \citep{li-etal-2020-unified}, because it uses additional external resources to construct the questions, which essentially introduces descriptive information about the categories.

\subsection{Detailed Parameter Settings}

In our experiments, the detailed parameter settings for the model are shown in Table \ref{tab:hp}.

\begin{table}[!t]
\small
\centering
\begingroup
\renewcommand{\arraystretch}{1.4} 
\begin{tabular}{c|c|c|c|c}
\toprule
\textbf{P}  & \textbf{ACE04}& \textbf{ACE05}& \textbf{KBP17}& \textbf{GENIA} \\
\midrule
lr & 3e-05 & 3e-05 & 5e-5 & 5e-6 \\
\hline
windows &  \multicolumn{3}{c|}{[1-7, 9, 11, 13, 15]} & [1-10] \\
\hline
batch size & 8 & 8 & 4 & 6 \\
\hline
$\gamma$& \multicolumn{4}{c}{2.0} \\
\hline
$\alpha_1$& \multicolumn{4}{c}{0.7} \\
\hline
$\alpha_2$& \multicolumn{4}{c}{1.0} \\
\hline
$\eta$& \multicolumn{4}{c}{1.0} \\
\hline
$u$& 0.9 & 0.8 & 0.9 &0.9  \\
\hline
$k$& 0.6 & 0.7 & 0.6 & 0.7  \\
\hline
$\delta$& 0.55 & 0.5 & 0.5 & 0.45\\
\hline
$\lambda_1, \lambda_2, \lambda_3$ & \multicolumn{4}{c}{[1.0, 0.1 ,1.0]}\\
\bottomrule
\end{tabular}
\endgroup
\caption{Detailed Parameter(P) Settings}
\label{tab:hp}
\end{table}

\subsection{Analysis of Boundary Offset Regression}

\begin{figure}[h]
  \centering
  \includegraphics[width=\linewidth]{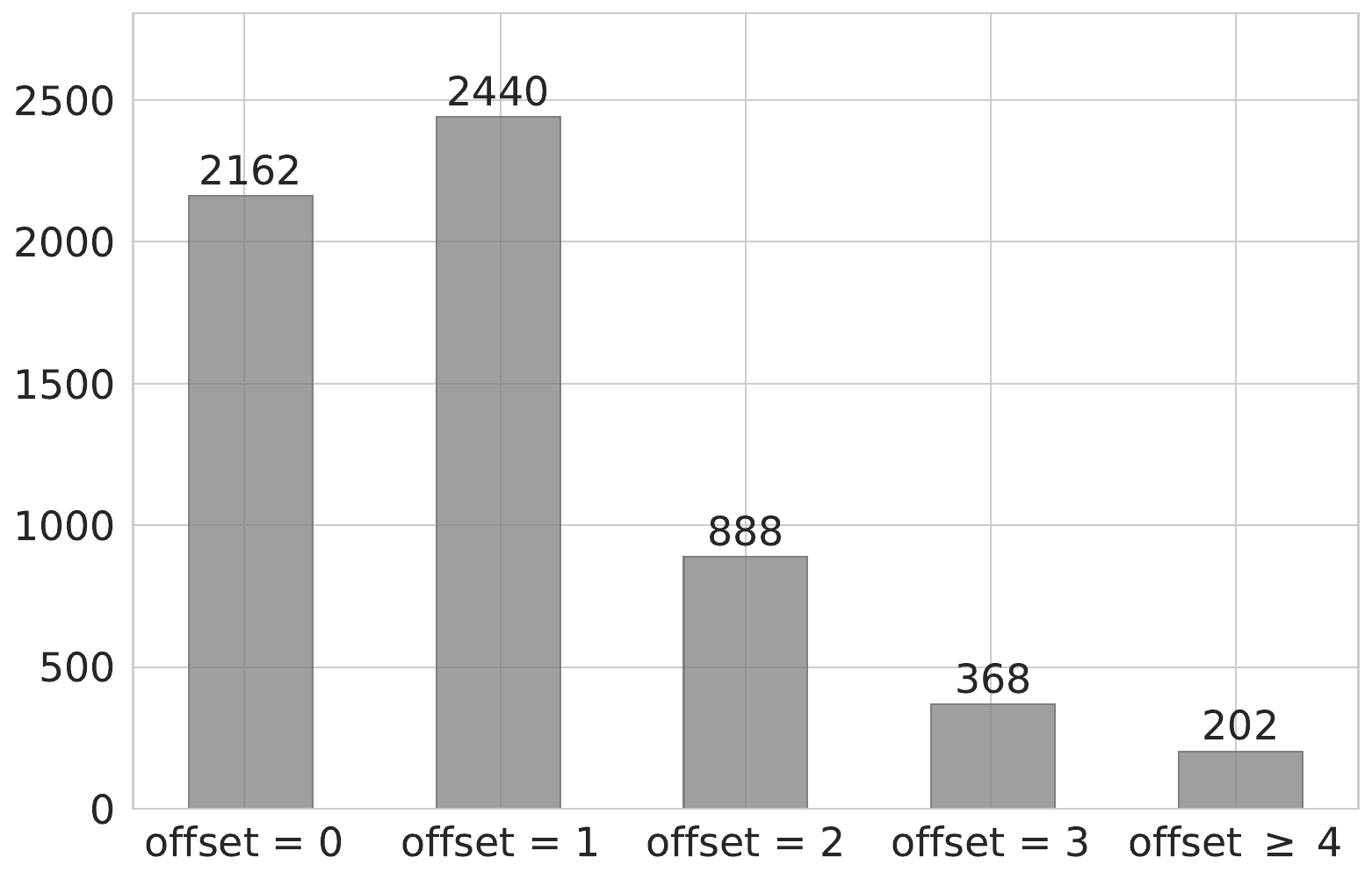}
  \caption{Boundary Offset Statistics}
  \label{fig:offset}
\end{figure}

We analyzed the distribution of the boundary offsets predicted by the model on the ACE04 dataset, as shown in Figure \ref{fig:offset}. We can find that the numbers of offsets by $0$, $1$, $2$, $3$ and $\geq4$  are $2162$, $2440$, $888$, $368$ and $202$, respectively. Most of the offsets are $1$, indicating that most of the seed spans require slight boundary adjustments to accurately locate the entities. There are also many offsets of $0$. This is because many entities in the dataset are short and the seed spans can cover them, and their boundaries do not need to be adjusted.

\section{Experiments on Flat NER}
\label{app:flatner}

\subsection{Datasets}

We use two flat NER datasets to evaluate our model:

\paragraph{CoNLL03 English} is an English dataset \citep{tjong-kim-sang-de-meulder-2003-introduction} with four types of flat entities: Location, Organization, Person and Miscellaneous. Following \citet{lin-etal-2019-sequence}, we train our model on the concatenation of the train and dev set.

\paragraph{Weibo Chinese} is a Chinese dataset \citep{peng-dredze-2015-named} sampled from Weibo with four types of flat entities, including Person, Organization, Location and Geo-political. And we evaluate our model using the same setting with \citet{li-etal-2020-flat}.

\subsection{Baselines}

For English flat NER, we use several taggers as baseline models, including  \textbf{ELMO-Tagger} \citep{peters-etal-2018-deep}, \textbf{BERT-Tagger} \citep{peters-etal-2018-deep}, which using ELMO, BERT as encoder respectively. And for Chinese flat NER, we use \textbf{Glyce} \citep{NEURIPS2019_452bf208}, \textbf{FLAT} \citep{li-etal-2020-flat} and \textbf{SLK-NER} \citep{hu2020slk} as baseline models. They incoprate glyph information, phrase embeddings and second-order lexicon knowledge for Chinese NER respectively.



\subsection{Results}

We evaluated our model on the flat NER dataset, as shown in Table \ref{tab:flat}. Our model outperforms the baseline models on Weibo Chinese, improving the F1-score by 0.61\%. On CoNLL03, our model also achieves comparable results, with less than 1\% performance drop compared to the \citep{yu-etal-2020-named}.

\begin{table}[t]
\small
\centering
\begin{tabular}{lccc}
\toprule
\multirow{2}{*}{Model}   & \multicolumn{3}{c}{CoNLL03 English}  \\
 \cmidrule(lr){2-4} 
& Pr.  & Rec. & F1  \\
\midrule

\citet{peters-etal-2018-deep} & -  & -  & 92.22 \\
\citet{devlin-etal-2019-bert} & -  & -  & 92.80 \\
\citet{yu-etal-2020-named} & 93.70  & 93.30  & 93.50     \\
\midrule
Ours    & 92.13  & \textbf{93.73}  & 92.94  \\
\bottomrule
\toprule
\multirow{2}{*}{Model}   & \multicolumn{3}{c}{Weibo Chinese}  \\
 \cmidrule(lr){2-4} 
& Pr.  & Rec. & F1  \\
\midrule

\citet{hu2020slk} & 61.80 &  66.30 &  64.00 \\
\citet{NEURIPS2019_452bf208} & 67.6 &  67.68 & 67.71 \\
\citet{li-etal-2020-flat} & -  & -  & 68.55 \\
\midrule
Ours    & \textbf{70.11}  & \textbf{68.12}  & \textbf{69.16}  \\
\bottomrule
\end{tabular}
\caption{Results for \textit{\textbf{flat}} NER tasks}
\label{tab:flat}
\end{table}

\end{document}